# Abstracting Probabilistic Actions

Peter Haddawy     AnHai Doan

Department of Electrical Engineering and Computer Science
University of Wisconsin-Milwaukee
PO Box 784
Milwaukee, WI 53201
{haddawy, anhai}@cs.uwm.edu

## Abstract

This paper discusses the problem of abstracting conditional probabilistic actions. We identify two distinct types of abstraction: intra-action abstraction and inter-action abstraction. We define what it means for the abstraction of an action to be correct and then derive two methods of intra-action abstraction and two methods of inter-action abstraction which are correct according to this criterion. We illustrate the developed techniques by applying them to actions described with the temporal action representation used in the DRIPS decision-theoretic planner and we describe how the planner uses abstraction to reduce the complexity of planning.

## 1 Introduction

Optimal planning in a decision-theoretic framework requires finding the plan or set of plans that maximizes expected utility. The complexity of decision-theoretic planning is a function of two factors: the number of possible plans and the length of plans. Complexity increases with the number of possible plans since searching for the optimal plan requires comparing the expected utilities of all possible plans. Complexity increases with the length of plans because computing the expected utility of a plan requires determining all possible outcomes of the plan. Since each action may have several possible outcomes and the total number of outcomes of a plan is the product of the number of outcomes of each of its actions, the number of outcomes of a plan increases exponentially as a function of plan length.

We discuss two types of abstraction for reducing the complexity of decision-theoretic planning. Inter-action abstraction can be used to reduce complexity as a function of the number of plans be grouping analogous actions together. Intra-action abstraction can reduce complexity as a function of plan length by grouping together different outcomes of individual actions.

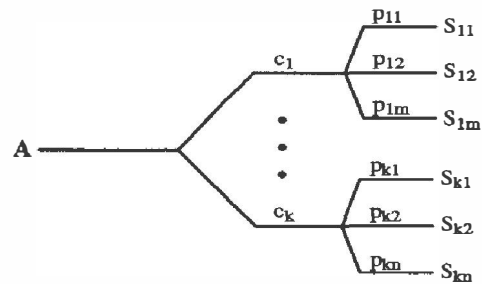

Figure 1: General form of an action description.

## 2 Preliminaries

In this section we discuss our representation of the world and of actions and we discuss the problem of projecting actions. Our representation is intentionally quite abstract to allow our results to be applied across different frameworks. We provide examples of applying the theory to a specific probabilistic temporal model in section 4.

### 2.1 Action and World Model

We represent the world with a propositional language. Since sentences are equivalent to sets of models, we will often use set notation when talking about sentences. Uncertainty is represented with a probability distribution over the models of the language. A model is simply a set of truth assignments to the propositional symbols and will also be referred to as a state.

Actions are both conditional and probabilistic: under certain conditions an action will have a given effect with a given probability. An action is represented with a tree structure as shown in figure 1, where the $c_i$ are a set of mutually exclusive and exhaustive conditions, the $p_{ij}$ are probabilities which sum to one for fixed $i$, and the $S_{ij}$ are effects. For each condition, the effects labeling the tips of the branches with that condition must be unique. If we had two identical effects following a condition, we could just combine the two branches and label it with the sum of the two proba-bilities.

We interpret the representation intuitively as saying that if one of the conditions holds at the beginning time of the action then the effects on that branch are realized immediately following the action, with the specified probabilities. Formally each branch represents a conditional probability statement. For example, the top branch in the figure means that $P(S_{11} \mid A \wedge c_1) = p_{11}$. We assume that the conditions labeling the branches are probabilistically independent of the action. This is reasonable since the conditions are conditions on the state of the world prior to performing the action. This representation is similar to that used by Hanks [Hanks, 1990b, Hanks, 1990a].

### 2.2 Projecting Actions

Decision-theoretic planning can be conceptually divided into two operations: generation of alternative plans and computation of the expected utilities of the generated plans. Computing the expected utility of a plan requires determining the outcomes of a plan, along with their probabilities. This we call plan projection. For simplicity of exposition, we will focus on the problem of projecting single actions. Projecting a single action can be cast as the the problem of computing the probability of any sentence conditioned on the action.

Suppose we want to compute the probability of a sentence $\phi$ given an action a. Since the conditions $c_i$ of the action are mutually exclusive and exhaustive,

$$P(\phi \mid \text{a}) = \sum_i P(\phi, c_i \mid \text{a})$$

By the assumption that actions are independent of their conditions,

$$P(\phi \mid \text{a}) = \sum_i P(\phi \mid c_i, \text{a}) \cdot P(c_i)$$

Since the $S_{ij}$ are mutually exclusive under each $c_i$ and their probabilities conditioned on $c_i$ sum to one, we have the following rule for projecting a concrete action

$$P(\phi \mid \text{a}) = \sum_{i,j} P(\phi \mid S_{ij}, c_i, \text{a}) \cdot P(S_{ij} \mid c_i, \text{a}) \cdot P(c_i)$$

Given some action descriptions, we are interested in generating sound descriptions of abstract actions. An abstraction is sound if the things we can infer from it are consistent with the things we can infer from its instances. Projecting an abstract action A will not in general produce a unique probability for any sentence $\phi$ but rather will provide a set of constraints on the probability of $\phi$. So an abstract action description is sound if the constraints it imposes on the probability of any sentence are consistent with those imposed by any of its instances.

**Definition 1 (Abstraction Criterion)** A *is an abstraction of an action* a *iff for any sentence $\phi$ the probability of $\phi$ resulting from projecting* a *is consistent with the constraints on the probability of $\phi$ resulting from projecting* A.

## 3 Abstracting Actions

Since abstract actions will in general impose only bounds on the probabilities of sentences, we need some way of referring to such bounds.

**Definition 2** *Let* S *be a set of states. We define the lower and upper probabilities of $\phi$ conditioned on* S *as*

$$P_*(\phi \mid \text{S}) = \begin{cases} 1 & \text{if } \text{S} \models \phi \\ 0 & \text{otherwise} \end{cases}$$

$$P^*(\phi \mid \text{S}) = \begin{cases} 0 & \text{if } \text{S} \models \neg\phi \\ 1 & \text{otherwise} \end{cases}$$

In producing projection rules for the various action abstractions, we will make use of the following theorem to compute probability bounds.

**Theorem 3** *For any set of states* S, $P_*(\phi \mid \text{S}) \leq P(\phi \mid \text{S}_i) \leq P^*(\phi \mid \text{S})$ *for all* $\text{S}_i \subseteq \text{S}$.

*Proof:* We provide the proof for $P_*$; the proof for $P^*$ is similar. If $\text{S} \models \phi$ then for all $\text{S}_i \subseteq \text{S}$, $\text{S}_i \models \phi$. So $P_*(\phi \mid \text{S}) = P(\phi \mid \text{S}_i) = 1$ for all $i$. If $\text{S} \not\models \phi$ then by definition $P_*(\phi \mid \text{S}) = 0$. So we have $P_*(\phi \mid \text{S}) \leq P(\phi \mid \text{S}_i)$ for all $\text{S}_i \subseteq \text{S}$.

### 3.1 Intra-Action Abstraction

We can reduce the branching factor of an action by abstracting its effects. The intra-action abstraction of an action description is an action description in which the branching factor has been reduced by replacing sets of branches of the original action with single branches. We derive two different methods of intra-action abstraction. The second method results in a more compact representation than the first but retains less of the information in the original action description. We derive the two abstraction methods by abstracting our projection rule for concrete actions.

We can uniquely specify the branch of an action by specifying the condition and the outcome. Suppose we group some of the branches of an action a into a set $\text{SC} = \langle S_{kl}, c_k \rangle$ then we can rewrite the projection rule as

$$P(\phi \mid \text{a}) =$$
$$\sum_{i,j : \langle S_{ij}, c_i \rangle \in \text{SC}} P(\phi \mid S_{ij}, c_i, \text{a}) \cdot P(S_{ij} \mid c_i, \text{a}) \cdot P(c_i) +$$
$$\sum_{i,j : \langle S_{ij}, c_i \rangle \notin \text{SC}} P(\phi \mid S_{ij}, c_i, \text{a}) \cdot P(S_{ij} \mid c_i, \text{a}) \cdot P(c_i)$$

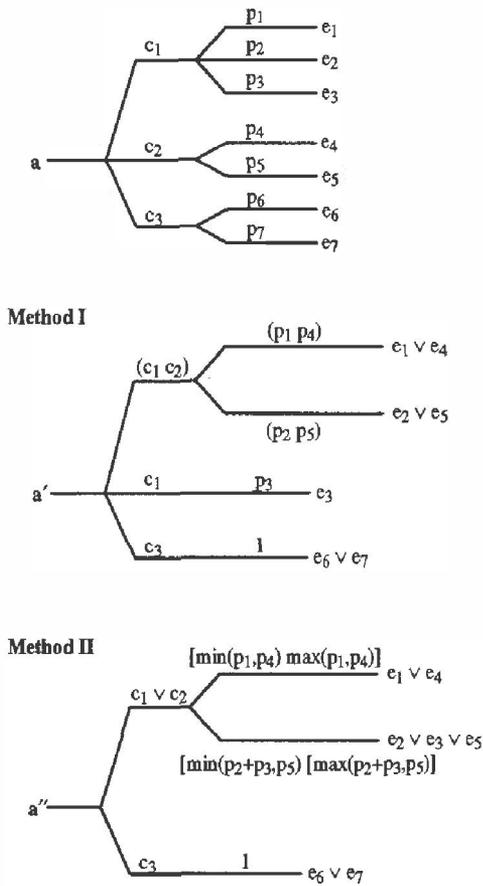

Figure 2: Two methods of intra-action abstraction.

Let $\mathbf{S}$ be any sentence such that $\mathbf{SC} \models \mathbf{S}$. Then by theorem (3) we can substitute $P_*(\phi \mid \mathbf{S}, \mathbf{a})$ for $P(\phi \mid S_{ij}, c_i, \mathbf{a})$ in the first term which results in the following projection rule

$P(\phi \mid \mathbf{a}) \geq$
$P_*(\phi \mid \mathbf{S}, \mathbf{a}) \cdot \sum_{i,j : <S_{ij}, c_i> \in \mathbf{SC}} P(S_{ij} \mid c_i, \mathbf{a}) \cdot P(c_i) +$
$\sum_{i,j : <S_{ij}, c_i> \notin \mathbf{SC}} P(\phi \mid S_{ij}, c_i, \mathbf{a}) \cdot P(S_{ij} \mid c_i, \mathbf{a}) \cdot P(c_i) \quad (1)$

The same holds if we change $P_*$ to $P^*$ and $\geq$ to $\leq$. This proof holds for grouping the branches into any number of disjoint sets. So we have the following intra-action abstraction method.

**Intra-Action Abstraction Method I**

Choose any way to group the branches of the action into disjoint sets. Produce an abstract action with one abstract branch for each set. The effect on an abstract branch is any sentence entailed by each of the effects in the set of grouped branches. The condition on an abstract branch is a list of the conditions on the branches in the set and the probability is a list of the probabilities corresponding to the conditions. When projecting the abstract action, for each branch we form the dot product of the vector of the probabilities of the conditions with the probability vector.

If the branches being grouped all share the same condition, say $c_1$, then we can produce a simpler description. In this case the new branch can just be labeled with that single condition and the sum of the probabilities.

**Theorem 4** *Intra-Action abstraction method I satisfies the abstraction criterion.*

*Proof:* This follows directly from the definition of the abstraction criterion and the derivation of equation 1.

Notice that according to this method a valid abstraction of an action is produced by simply weakening some of the action's effects. In other words, we could group the branches into singleton sets and replace any of the effects with a sentence entailed by the effect.

Figure 2 shows an application of inter-action abstraction method I. Abstract action description $a'$ is obtained from action description $a$ by grouping the branches $(e_1, e_4)$, $(e_2, e_5)$, and $(e_6, e_7)$.

Suppose that we wish to obtain greater compression of the information in the action description than with our above representation. A lower bound on the summation in the first term of inequality 1 is

$\min_{j \,:\, \langle S_{ij}, c_i \rangle \in \mathbf{SC}} P(\phi \mid S_{ij}, c_i, \mathbf{a}) \cdot \sum_{i \,:\, \langle S_{ij}, c_i \rangle \in \mathbf{SC}} P(c_i)$

But since the conditions of an action are mutually exclusive, this is just

$\min_{j \,:\, \langle S_{ij}, c_i \rangle \in \mathbf{SC}} P(\phi \mid S_{ij}, c_i, \mathbf{a}) \cdot$
$P(\bigvee_{i \,:\, \langle S_{ij}, c_i \rangle \in \mathbf{SC}} c_i)$

So we have
$P(\phi \mid \mathbf{a}) \geq$
$P_*(\phi \mid \mathbf{S}, \mathbf{a}) \cdot \min_{j \,:\, \langle S_{ij}, c_i \rangle \in \mathbf{SC}} P(\phi \mid S_{ij}, c_i, \mathbf{a}) \cdot$
$P(\bigvee_{i \,:\, \langle S_{ij}, c_i \rangle \in \mathbf{SC}} c_i) +$
$\sum_{i,j : <S_{ij}, c_i> \notin \mathbf{SC}} P(\phi \mid S_{ij}, c_i, \mathbf{a}) \cdot P(S_{ij} \mid c_i, \mathbf{a}) \cdot P(c_i) \quad (2)$

and we can specify an upper bound by replacing min with max. This gives us our second intra-action abstraction method.

**Intra-Action Abstraction Method II**

Choose any way to group the branches of the action into disjoint sets. Produce an abstract action with one abstract branch for each set. The effect on an abstract branch is any sentence entailed by each of the effects in the set of grouped branches. The condition on an abstract branch is the disjunction of the conditions on the branches in the set and the probability is a range whose lower and upper bounds are the minimum and maximum of the probabilities on the branches in the set, respectively. When projecting the abstract action, for each branch we compute the range of probability values resulting from the product of the probability of the condition and the two probability bounds.

**Theorem 5** *Intra-Action abstraction method II satisfies the abstraction criterion.*

*Proof:* This follows directly from the definition of the abstraction criterion and the derivation of equation 2.

An application of abstraction method II is shown in at the bottom right of figure 2. Action a has been abstracted by first grouping branches $e_2$ and $e_3$ and then grouping the branches $(e_1, e_4)$, $((e_2, e_3), e_5)$, and $(e_6, e_7)$.

### 3.2  Inter-Action Abstraction

We extend Tenenberg's [Tenenberg, 1991] notion of inheritance abstraction for STRIPS operators to apply to conditional probabilistic actions. As Tenenberg explains it, "the intent of using inheritance abstraction is to formalize the notion that analogous action types can be structured together into an action class at the abstract level characterized by the common features of all elements of the class." Thus we can plan with the abstract action and infer properties of a plan involving any of the instances of the abstract action.

**Definition 6 (Inter-Action Abstraction)**
*An inter-action abstraction of a set of actions $\{a^1, a^2, ...a^n\}$ is an action that represents the disjunction of the actions in the set. The actions in the set are called the* instantiations *of the abstract action and are considered to be alternative ways of realizing the abstract action. Thus the $a^i$ are assumed to be mutually exclusive.*

As in the previous section, we derive two methods for inter-action abstraction. The second method results in a more abstract representation than the first. We will show how to create a description of the inter-action abstraction of a set of actions by first deriving a projection rule for the abstract action. The rule will lead directly to an abstraction method that satisfies the abstraction criterion.

Suppose we wish to abstract a pair of actions $a^1$ and $a^2$. The generalization to an arbitrary number of actions will be obvious. We can project each action according to the projection rule:

$$P(\phi \mid a^1) = \sum_{i,j} P(\phi \mid S_{ij}, c_i, a^1) \cdot P(S_{ij} \mid c_i, a^1) \cdot P(c_i)$$

$$P(\phi \mid a^2) = \sum_{i,j} P(\phi \mid S_{ij}, c_i, a^2) \cdot P(S_{ij} \mid c_i, a^2) \cdot P(c_i)$$

Since $a^1$ and $a^2$ are mutually exclusive,

$$\min[P(\phi \mid a^1), P(\phi \mid a^2)] \leq P(\phi \mid a^1 \vee a^2) \leq$$

$$\max[P(\phi \mid a^1), P(\phi \mid a^2)]$$

We wish to derive a projection rule for the abstract action by pairing the branches of the two actions. But the two actions may not have the same number of branches. We can remedy the situation by taking the action description with fewer branches and adding branches with arbitrary effects, condition False, and probability zero. When projecting the action the outcomes specified by those branches will be assigned probability zero, which is exactly what the original action description is doing implicitly by not containing those branches. So assume that $a^1$ and $a^2$ have the same number of branches and pair the branches in any way so that we can write

$$\sum_{ij} \min[P(\phi \mid S^1_{ij}, c^1_i, a^1) \cdot P(S^1_{ij} \mid c^1_i, a^1) \cdot P(c^1_i),$$
$$P(\phi \mid S^2_{ij}, c^2_i, a^2) \cdot P(S^2_{ij} \mid c^2_i, a^2) \cdot P(c^2_i)]$$
$$\leq P(\phi \mid a^1 \vee a^2) \leq$$
$$\sum_{ij} \max[P(\phi \mid S^1_{ij}, c^1_i, a^1) \cdot P(S^1_{ij} \mid c^1_i, a^1) \cdot P(c^1_i),$$
$$P(\phi \mid S^2_{ij}, c^2_i, a^2) \cdot P(S^2_{ij} \mid c^2_i, a^2) \cdot P(c^2_i)]$$

For each pair of branches, $(S^1_{ij}, S^2_{ij})$, let $\mathbf{S}_{ij}$ be any sentence such that $S^1_{ij} \models \mathbf{S}_{ij}$ and $S^2_{ij} \models \mathbf{S}_{ij}$. Then by theorem (3) we have

$$\sum_{ij} P_*(\phi \mid \mathbf{S}_{ij}, c^1_i, a^1) \cdot$$
$$\min[P(S^1_{ij} \mid c^1_i, a^1) \cdot P(c^1_i),\ P(S^2_{ij} \mid c^2_i, a^2) \cdot P(c^2_i)]$$
$$\leq P(\phi \mid a^1 \vee a^2) \leq \qquad (3)$$
$$\sum_{ij} P^*(\phi \mid \mathbf{S}_{ij}, c^1_i, a^1) \cdot$$
$$\max[P(S^1_{ij} \mid c^1_i, a^1) \cdot P(c^1_i),\ P(S^2_{ij} \mid c^2_i, a^2) \cdot P(c^2_i)]$$

**Inter-Action Abstraction Method I**

To create an abstraction of a set of actions $\{a_1, a_2, ..., a_n\}$ we do the following. Group the branches of the action descriptions into disjoint sets such that each set contains at most one branch from each action description. For each set $s$ that contains fewer than $n$ branches, add $n - |s|$ branches, each with the effect of one of the branches already in the set and with condition False and probability zero. The effect of an abstract branch is any sentence entailed by each of the effects of the branches in the set. The condition and probability are lists of the conditions and the corresponding probabilities of the branches in the set.

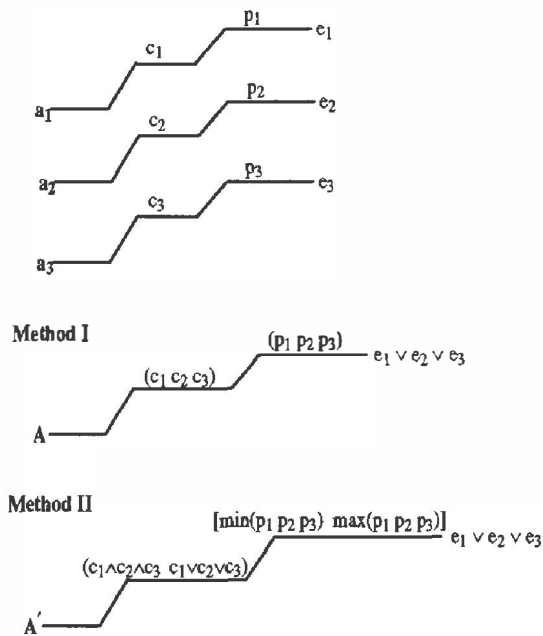

Figure 3: Two methods of inter-action abstraction.

**Theorem 7** *Inter-Action abstraction method I satisfies the abstraction criterion.*

*Proof:* This follows directly from the definition of the abstraction criterion and the derivation of inequality 3.

Figure 3 shows an application of inter-action abstraction method I. At the top of the figure is shown one branch of the action description for each of three actions. Under the label "Method I" is the abstract branch that results from grouping these.

The problem with keeping lists of conditions and probabilities is that the lists may grow very long. In an abstraction hierarchy of depth $k$ in which each abstract action description has $n$ instances, the length of the condition and probability lists on the top-level abstract action would be $n^k$. But we would like to use deep abstraction hierarchies if possible because for a given set of concrete actions, the deeper the hier-archy generally the more effectively we can prune the space of possible plans. We can avoid the problem of increasingly long lists of conditions and probabilities by further abstracting the action description. Since $P(c_i^1 \wedge c_i^2) \leq P(c_i^k)$, $k = 1, 2$ and $P(c_i^1 \vee c_i^2) \geq P(c_i^k)$, $k = 1, 2$, inequality 3 may be written as

$$\begin{aligned}
&\sum_{ij} P_*(\phi \mid \mathbf{S}_{ij}, c_i^1, \mathbf{a}^1) \cdot P(c_i^1 \wedge c_i^2) \cdot \\
&\quad \min[P(S_{ij}^1 \mid c_i^1, \mathbf{a}^1), \ P(S_{ij}^2 \mid c_i^2, \mathbf{a}^2)] \\
&\leq P(\phi \mid \mathbf{a}^1 \vee \mathbf{a}^2) \leq \\
&\sum_{ij} P^*(\phi \mid \mathbf{S}_{ij}, c_i^1, \mathbf{a}^1) \cdot P(c_i^1 \vee c_i^2) \cdot \\
&\quad \max[P(S_{ij}^1 \mid c_i^1, \mathbf{a}^1), \ P(S_{ij}^2 \mid c_i^2, \mathbf{a}^2)]
\end{aligned} \quad (4)$$

**Inter-Action Abstraction Method II**

To create an abstraction of a set of actions $\{a_1, a_2, ..., a_n\}$ we group the branches of the action descriptions into disjoint sets as in Method I. The effect of an abstract branch is any sentence entailed by each of the effects of the branches in the set. The condition is specified by two formulas: the conjunction of the conditions on the branches in the set and the disjunction of the conditions on the branches in the set. The probability is specified as a range: the minimum of the probabilities of the branches in the set and the maximum of the probabilities of the branches in the set.

**Theorem 8** *Inter-Action abstraction method II satisfies the abstraction criterion.*

*Proof:* This follows directly from the definition of the abstraction criterion and the derivation of inequality 4.

At the bottom of figure 3 is shown the action description that results from applying inter-action abstraction method II to the three branches at the top of the figure. The condition on the branch of the resulting abstract action can be thought of as specifying a range of models since the set of models satisfying the conjunction is a strict subset of the set of models satisfying the disjunction. The disjunction specifies the conditions necessary for the effects to be realized and the conjunction specifies the sufficient conditions. If we wish to further abstract the action description, we can replace the conjunction with any sentence that entails it and we can replace the disjunction with any sentence entailed by it.

## 4   Applying the Techniques

In this section we present some examples of applying the abstraction methods to actions described in

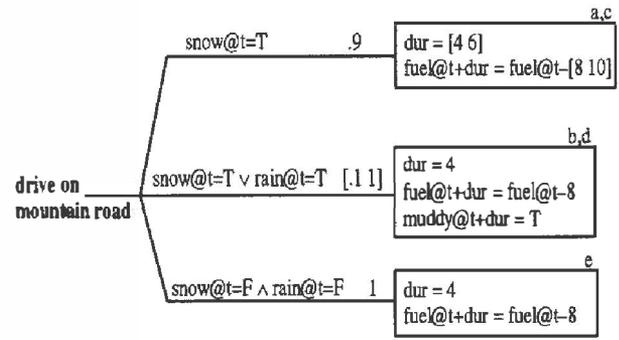

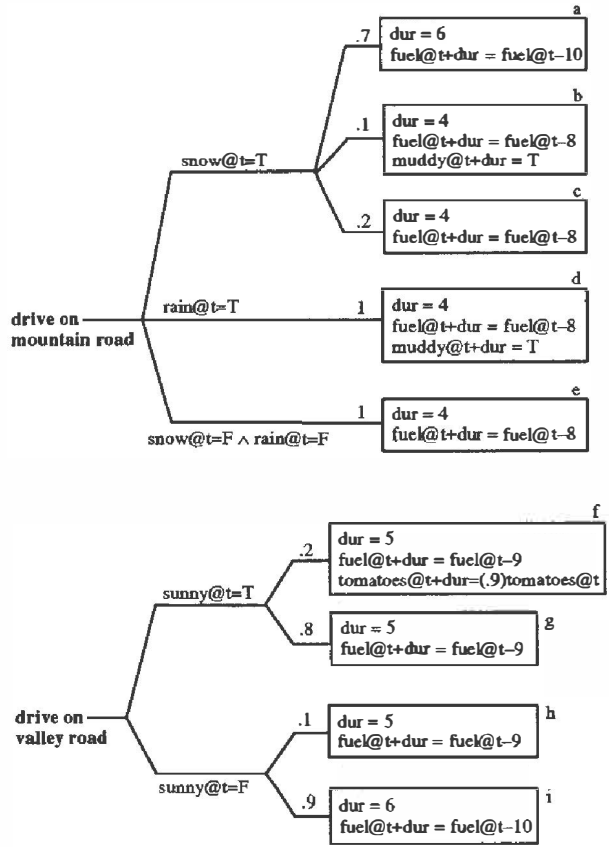

Figure 4: Example action descriptions.

Figure 5: Example of intra-action abstraction.

Using the probability operator we can express uncertainty about the state of the world. For example, to say that there is a 70% chance that we will use eight gallons of fuel driving home from time $t_0$ to time $t_1$ we might write

$$P(\text{fuel}@t_1 = \text{fuel}@t_0 - 8 \mid \text{drive-home}@t_0 = T) = .7$$

We represent the state of the world at any given time with a probability distribution over a finite set of fluents. For a complete development of a more expressive fully first-order temporal probability logic see [Haddawy, 1994].

### 4.2  Examples

To illustrate the action representation, consider the following example. We wish to deliver some tomatoes and we can take a mountain road or a valley road to make the delivery. If we take the mountain road, the length of time the drive takes, the fuel consumption, and whether the truck gets muddy depend on the weather conditions. If we take the valley road and the sun is shining, there is a 20% chance that 10% of the tomatoes will spoil. The descriptions of these two actions are shown in figure 4. The time constant *dur* specifies the duration of the action. We make the traditional STRIPS assumption that the only changes to the world are those explicitly mentioned in the action descriptions. Since the state of the world is unchanged unless explicitly changed by an action, the world remains in its state prior to the action until the time $t + dur$. Action effects can be specified as absolute or relative. In branch $b$ we have the absolute effect muddy@$t$=T, and the relative effect fuel@$t + dur$ = fuel@$t - 8$. So branch $b$ says that if it is snowing just prior to driving on the mountain road then there is a 10% chance that the drive will require 4 hours, and that the truck will consume 8 gallons of fuel and become muddy.

Figure 5 shows an abstraction of the "drive on mountain road" action using intra-action abstraction method II. The action was abstracted by grouping the representation language of the DRIPS decision-theoretic refinement planning system [Haddawy and Suwandi, 1994]. The DRIPS system uses abstraction to reduce the complexity of planning by eliminating suboptimal abstract plans.

### 4.1  The DRIPS Representation

The DRIPS system reasons with a probabilistic temporal world model. The lexicon of the language contains only the following non-logical symbols: fluent symbols, object constant symbols, time constant symbols, temporal variables, numeric constant symbols, numeric function symbols, and numeric relation symbols. The language includes the usual logical operators and quantifiers, as well as the probability operator $P$. A fluent is a function from time points to a range of values. If $F$ is a fluent we will write $F@t$ to designate the value of $F$ at time $t$. So, for example, we could write fuel@$t_0$=10 to say that the fuel level at time $t_0$ is 10 gallons. Notice that the language contains no predicates other than the numeric relations. So we represent attributes that would normally be represented as predicates with boolean-valued functions. For example, to say that the truck is muddy at time $t_0$ we could write muddy@$t_0$=T.

the branches into the sets $(a, c)$ and $(b, d)$. Notice that rather than specifying the effects of the abstract branches as the disjunction of the effects of the grouped branches we have represented the effects as ranges of values. The ability to arbitrarily weaken effects provides us with important representational economy when working with continuous attributes.

Figure 6 show the abstraction of the two drive actions into the action "drive" using inter-action abstraction method I. The outcomes of the abstract action are labeled with the branches that were grouped to form them. For example, the top outcome of the drive action was created by grouping branch $a$ of "drive on mountain road" and branch $i$ of "drive on valley road". Notice that branch $b$ of "drive on mountain road" was not grouped with any other branch, so in producing the abstract action description, it was grouped with a branch of probability zero and condition False of the "drive on valley road" action. Thus the abstract action description has the same number of branches as the instance with the greatest number of branches. Notice also that outcome $d, g$ of "drive" has a range for muddy that includes muddy@t. The reason for this is that outcome $d$ changes the value of muddy to T while outcome $g$ leaves it unchanged by the STRIPS assumption. When we group the two outcomes, we need to indicate that the value is either T or unchanged.

### 4.3 The DRIPS Algorithm

The DRIPS system uses inter-action abstraction method II to reduce the complexity of planning. The planner reasons with action descriptions organized into an abstraction/decomposition network. Abstractions indicate various choices for instantiating abstract actions and decompositions indicate how tasks are decomposed into subtasks. Given an abstraction/decomposition network, a probability distribution over initial world states, and a utility function over chronicles, the planner searches for the plan that maximizes expected utility by building abstract plans, comparing their expected utilities, and refining only those abstract plans that might be refinable to the optimal plan. Since projecting abstract actions results in inferring probability intervals and attribute ranges, abstract plans will be assigned expected utility intervals. An abstract plan can be eliminated if the upper bound of its expected utility interval is lower than the lower bound of the expected utility interval for another plan. When the planner has some abstract plans with overlapping expected utility intervals, it refines the plans by instantiating some of their actions. This causes the expected utility intervals to narrow. If some of the new intervals do not overlap, more plans can be eliminated and the planner can then refine those remaining one step further.

The planner's use of abstraction can result in a large computational savings, which becomes more significant as the size of the planning problem increases. Suppose we have an abstraction/decomposition network with $p$ actions in each decomposition, $n$ possible instantiations of each abstract action, and $k$ levels of abstraction. This network contains $n^{(p+p^2+p^3+\ldots+p^k)}$ possible concrete plans. Now suppose that $x$ is the percentage of the plans remaining at each abstraction level after pruning. The number of plans (abstract and concrete) for which the DRIPS algorithm computes expected utility is

$$n + xn^2 + x^2n^3 + \ldots + x^{p+p^2+\ldots+p^k-1}n^{p+p^2+\ldots+p^k}.$$

With maximum pruning $(xn = 1)$ the number of plans examined is only

$$n + xn \times n + (xn)^2 \times n + \ldots + (xn)^{p+p^2+\ldots+p^k-1} \times n =$$

$$n(p + p^2 + \ldots + p^k),$$

a logarithmic reduction in complexity.

## 5 Related Work

Hanks [1990a] presents a method for reducing the complexity of projection similar in concept to our intra-action abstraction technique. He examines the problem of projecting actions to determine the probability of a goal proposition. When the projection of an action results in outcomes that are distinguished according to propositions irrelevant to the goal, those outcomes are "bundled" together to produce an outcome set. Since his technique is designed for abstracting outcomes on the fly, he does not address abstracting action descriptions.

Our method of representing abstract actions is similar to that of Chrisman [1992]. He represents the uncertainty in the effects of an abstract action by using Dempster-Shafer intervals. He derives a closed-form projection rule that works by finding the convex hull of possible poststate probability distributions. Although his actions can include conditions, he does not show how to abstract the conditions.

## 6 Current and Future Research

We have presented several sound methods for abstracting probabilistic actions which are useful for reducing the complexity of planning. These methods are like inference rules that can be applied to action descriptions in any number of ways. For both types of abstraction, we must decide which branches to group and in the case of inter-action abstraction we must additionally decide which actions to group. These decisions are important since an abstraction will be effective only if the resulting probability intervals are not too wide and if the abstraction is well tailored to the utility function for which a plan is being generated. Intuitively, we would like to abstract away unimportant attributes and retain detailed information concerning those that have a large influence on utility. We are currently investigating methods for automatically grouping actions into abstraction hierarchies based on information concerning the form of the utility function.

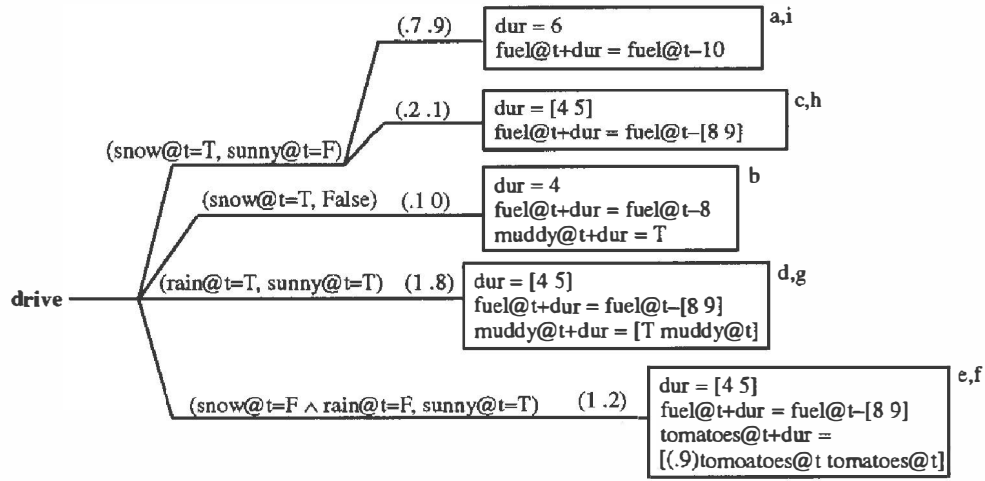

Figure 6: Example of inter-action abstraction.

We need yet to determine the effectiveness of the abstraction techniques on large practical problems. We are currently applying the DRIPS planner to the problem of generating plans for brewing beer and to the problem of choosing strategies for the diagnosis and treatment of lower-extremity deep vein thrombosis. These are both large domains containing several thousand plans. Thus far the abstraction techniques have been relatively easy to apply and the planner has performed well. Appropriate abstraction hierarchies have naturally arisen out of the problem descriptions and the planner has been able to efficiently find the optimal plan. In the brewing domain the planner is able to find the optimal plan after only examining between 4% and 25% of the plans represented in the network. The percentage depends on the type of beer being planned for, which is specified in the utility function.

We have developed our abstraction methods for actions with discrete outcomes. When action effects are specified in terms of continuous attributes, the probabilities are often more accurately described in terms of continuous distributions over those attributes. We are currently working on extending our abstraction techniques to encompass actions with effects described as continuous distributions.

The DRIPS planner currently uses only inter-action abstraction, so it addresses complexity as a function of the number of plans. Incorporating intra-action abstraction would reduce the complexity as a function of plan length. Integrating the two kinds of abstraction raises interesting control issues since at any point where we can refine a plan we can choose to refine along either of the two abstraction dimensions.

### Acknowledgements

This work was partially supported by NSF grant #IRI-9207262.